\documentclass[runningheads]{llncs}
\usepackage{graphicx}

\usepackage{tikz}
\usepackage{comment} 
\usepackage{amsmath,amssymb} 
\usepackage{color}

\usepackage[width=122mm,left=12mm,paperwidth=146mm,height=193mm,top=12mm,paperheight=217mm]{geometry}

\usepackage{hyperref}
\usepackage{subfigure}
\usepackage{booktabs}
\usepackage{times}
\usepackage{epsfig}
\DeclareMathOperator*{\argmax}{argmax}

\begin{document}
\pagestyle{headings}
\mainmatter
\def\ECCVSubNumber{4584}  

\title{Ultra Fast Structure-aware Deep Lane Detection \thanks{\scriptsize This work is supported by key scientific technological innovation research project by Ministry of Education, Zhejiang Provincial Natural Science Foundation of China under
Grant LR19F020004, Baidu AI Frontier Technology Joint Research Program, and Zhejiang University K.P.Chao's High Technology Development Foundation. }} 

\titlerunning{Ultra Fast Structure-aware Deep Lane Detection}
%
\author{Zequn Qin \and
Huanyu Wang \and
Xi Li\thanks{Corresponding author.}\orcidID{0000-0003-3023-1662}}
\authorrunning{Z. Qin et al.}
%
\institute{College of Computer Science \& Technology, \\Zhejiang University, Hangzhou, China\\
\email{zequnqin@gmail.com, \{huanyuhello,xilizju\}@zju.edu.cn}}

\maketitle
\begin{abstract}
Modern methods mainly regard lane detection as a problem of pixel-wise segmentation, 
which is struggling to address the problem of challenging scenarios and speed.
Inspired by human perception, the recognition of lanes under severe occlusion and extreme lighting conditions is mainly based on contextual and global information.
Motivated by this observation, we propose a novel, simple, yet effective formulation aiming at extremely fast speed and challenging scenarios. Specifically, we treat the process of lane detection as a row-based selecting problem using global features. With the help of row-based selecting, our formulation could significantly reduce the computational cost. Using a large receptive field on global features, we could also handle the challenging scenarios. Moreover, based on the formulation, we also propose a structural loss to explicitly model the structure of lanes.
Extensive experiments on two lane detection benchmark datasets show that our method could achieve the state-of-the-art performance in terms of both speed and accuracy. A light weight version could even achieve 300+ frames per second with the same resolution, which is at least 4x faster than previous state-of-the-art methods. Our code is available at \url{https://github.com/cfzd/Ultra-Fast-Lane-Detection}.

\end{abstract}

\section{Introduction}
With a long research history in computer vision, lane detection is a fundamental problem and has a wide range of applications \cite{hillel2014recent} (e.g., ADAS and autonomous driving). 
For lane detection, there are two kinds of mainstream methods, which are traditional image processing methods \cite{bertozzi1998gold,wang2004lane,aly2008real} and deep segmentation methods  \cite{Empirical_Evaluation,SCNN,End-to-End}. Recently, deep segmentation methods have made great success in this field because of great representation and learning ability. There are still some important and challenging problems to be addressed.

As a fundamental component of autonomous driving, the lane detection algorithm is heavily executed. This requires an extremely low computational cost of lane detection. Besides, present autonomous driving solutions are commonly equipped with multiple camera inputs, which typically demand lower computational cost for every camera input. In this way, a faster pipeline is essential to lane detection. For this purpose, SAD \cite{SAD} is proposed to solve this problem by self-distilling. Due to the dense prediction property of SAD, which is based on segmentation, the method is computationally expensive. 

\begin{figure}[t]
	\centering
	\includegraphics[width=\columnwidth]{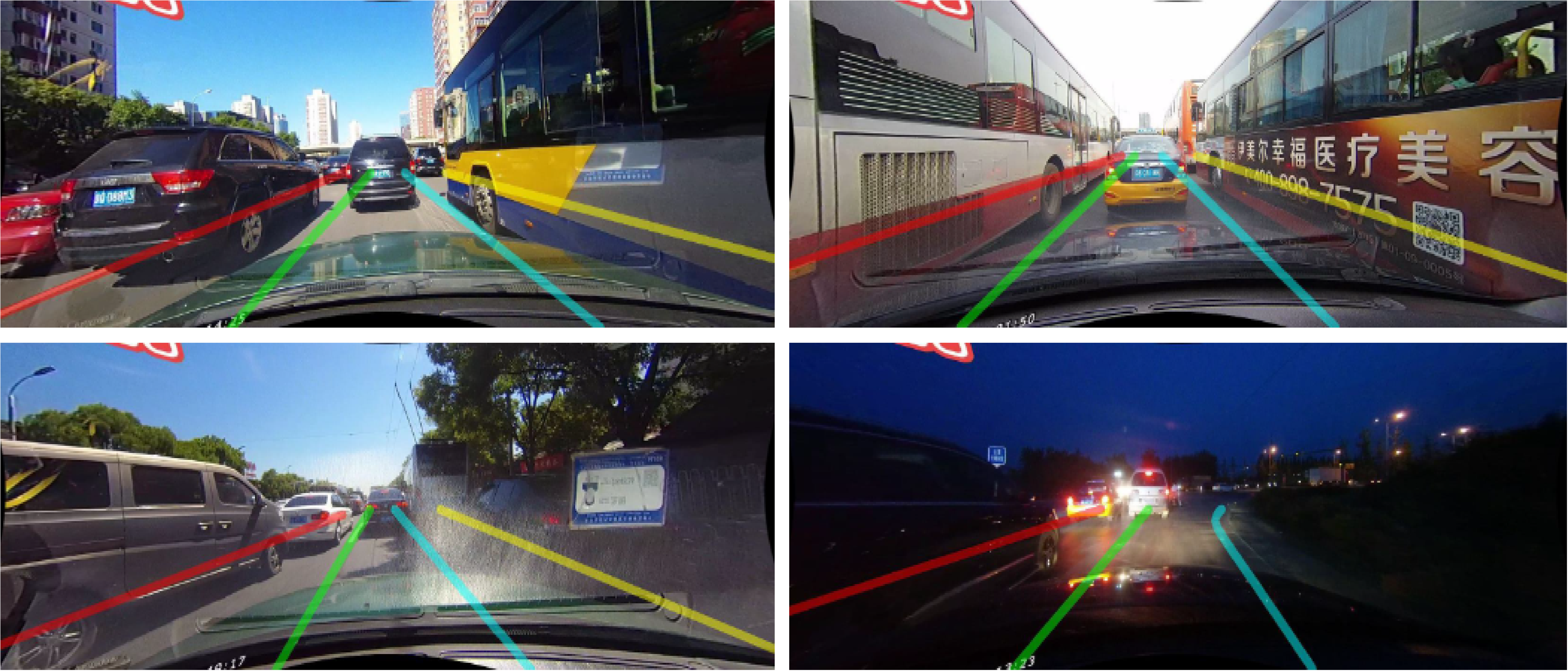}
	\caption{Illustration of difficulties in lane detection. Different lanes are marked with different colors. Most of challenging scenarios are severely occluded or distorted with various lighting conditions, resulting in little or no visual clues of lanes can be used for lane detection.}
	\label{figure_intro}
	\vspace{-20pt}
\end{figure}

Another problem of lane detection is called \textit{no-visual-clue}, as shown in Fig. \ref{figure_intro}. Challenging scenarios with severe occlusion and extreme lighting conditions correspond to another key problem of lane detection. In this case, the lane detection urgently needs higher-level semantic analysis of lanes. Deep segmentation methods naturally have stronger semantic representation ability than conventional image processing methods, and become mainstream. Furthermore, SCNN \cite{SCNN} addresses this problem by proposing a message passing mechanism between adjacent pixels, which significantly improves the performance of deep segmentation methods. Due to the dense pixel-wise communication, this kind of message passing requires a even more computational cost.

 Also, there exists a phenomenon that the lanes are represented as segmented binary features rather than lines or curves. Although deep segmentation methods dominate the lane detection fields, this kind of representation makes it difficult for these methods to explicitly utilize the prior information like rigidity and smoothness of lanes.

With the above motivations, we propose a novel lane detection formulation aiming at extremely fast speed and solving the \textit{no-visual-clue} problem. Meanwhile, based on the proposed formulation, we present a structural loss to explicitly utilize prior information of lanes.
Specifically, our formulation is proposed to \textbf{select locations of lanes at predefined rows of the image using global features} instead of segmenting every pixel of lanes based on a local receptive field, which significantly reduces the computational cost. The illustration of location selecting is shown in Fig. \ref{fig_formulation_new}.

For the \textit{no-visual-clue} problem, our method could also achieve good performance, because our formulation is conducting the procedure of selecting in rows based on global features. With the aid of global features, our method has a receptive field of the whole image. Compared with segmentation based on a limited receptive field, visual clues and messages from different locations can be learned and utilized. In this way, our new formulation could solve the speed and the \textit{no-visual-clue} problems simultaneously. Moreover, based on the proposed formulation, lanes are represented as selected locations on different rows instead of the segmentation map. Hence, we can directly utilize the properties of lanes like rigidity and smoothness by optimizing the relations of selected locations, i.e., the structural loss. The contribution of this work can be summarized in three parts:

\begin{figure*}[t]
	\centering
	\includegraphics[width=0.95\linewidth]{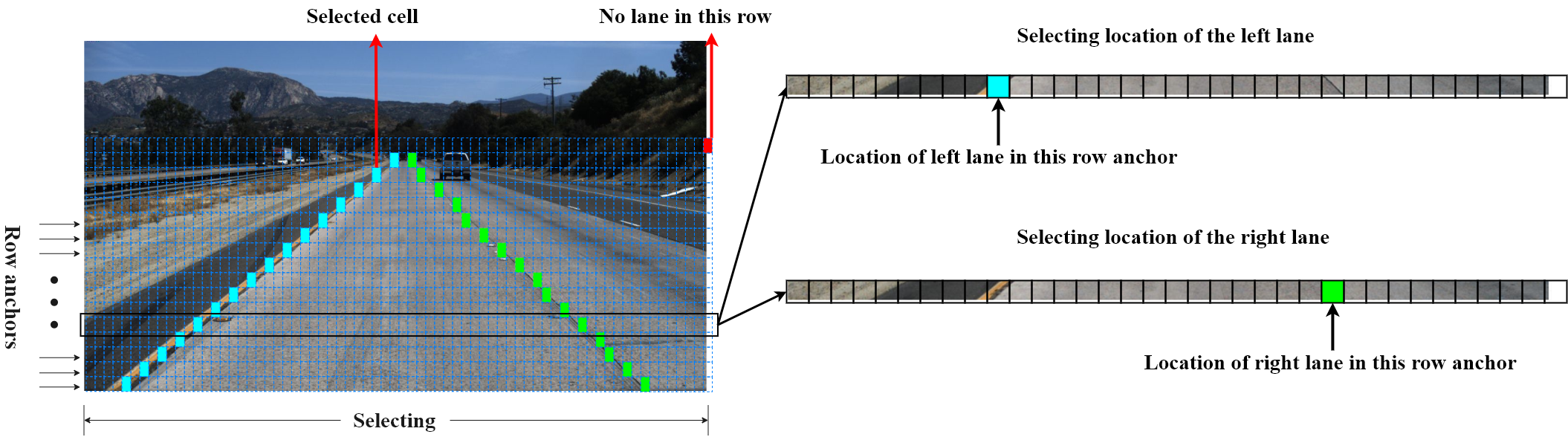}
	\caption{Illustration of selecting on the left and right lane. In the right part, the selecting of a row is shown in detail. Row anchors are the predefined row locations, and our formulation is defined as horizontally selecting on each of row anchor. On the right of the image, a background gridding cell is introduced to indicate no lane in this row.}
	\label{fig_formulation_new}
\vspace{-10pt}
\end{figure*}

\begin{itemize}
	\item We propose a novel, simple, yet effective formulation of lane detection aiming at extremely fast speed and solving the \textit{no-visual-clue} problem. Compared with deep segmentation methods, our method is selecting locations of lanes instead of segmenting every pixel and works on the different dimensions, which is ultra fast. Besides, our method uses global features to predict, which has a larger receptive field than the segmentation formulation. In this way, the \textit{no-visual-clue} problem can also be addressed.
	\item Based on the proposed formulation, we present a structural loss which explicitly utilizes prior information of lanes. To the best of our knowledge, this is the first attempt at optimizing such information explicitly in deep lane detection methods.
	\item The proposed method achieves the state-of-the-art performance  in terms of both accuracy and speed on the challenging CULane dataset. A light weight version of our method could even achieve 300+ FPS with a comparable performance with the same resolution, which is at least 4 times faster than previous state-of-the-art methods.

\end{itemize}

\section{Related Work}
\subsubsection{Traditional methods}
Traditional approaches usually solve the lane detection problem based on visual information. The main idea of these methods is to take advantage of visual clues through image processing like the HSI color model \cite{sun2006hsi} and edge extraction algorithms \cite{yu1997lane,wang2000lane}. When the visual information is not strong enough, tracking is another popular post-processing solution \cite{wang2004lane,kim2008robust}. Besides tracking, Markov and conditional random fields \cite{CRF} are also used as post-processing methods. With the development of machine learning, some methods \cite{kluge1995deformable,gonzalez2000lane,mandalia2005using} that adopt algorithms like template matching and support vector machines are proposed. 

\subsubsection{Deep learning Models}
With the development of deep learning, some methods \cite{kim2014robust,Empirical_Evaluation} based on deep neural networks show the superiority in lane detection. These methods usually use the same formulation by treating the problem as a semantic segmentation task. For instance, VPGNet \cite{Lee_2017_ICCV} proposes a multi-task network guided by vanishing points for lane and road marking detection. To use visual information more efficiently, SCNN \cite{SCNN} utilizes a special convolution operation in the segmentation module. It aggregates information from different dimensions via processing sliced features and adding them together one by one, which is similar to the recurrent neural networks. Some works try to explore light weight methods for real-time applications. Self-attention distillation (SAD) \cite{SAD} is one of them. It applies an attention distillation mechanism, in which high and low layers' attentions are treated as teachers and students, respectively. 

Besides the mainstream segmentation formulation, other formulations like Sequential prediction and clustering are also proposed. In \cite{li2016deep}, a long short-term memory (LSTM) network is adopted to deal with the long line structure of lanes. With the same principle, Fast-Draw \cite{FastDraw} predicts the direction of lanes at each lane point, and draws them out sequentially. In \cite{Proposal-Free}, the problem of lane detection is regarded as clustering binary segments. The method proposed in \cite{agnostic} also uses a clustering approach to detect lanes. Different from the 2D view of previous works, a lane detection method in 3D formulation \cite{garnett20193d} is proposed to solve the problem of non-flatten ground.

\section{Method}
In this section, we describe the details of our method, including the new formulation and lane structural losses. Besides, a feature aggregation method for high-level semantics and low-level visual information is also depicted.
\vspace{-10pt}
\subsection{New formulation for lane detection}
\label{sec_formulation}
As described in the introduction section, fast speed and the \textit{no-visual-clue} problems are important for lane detection. Hence, how to effectively handle these problems is key to good performance. In this section, we show the derivation of our formulation by tackling the speed and the \textit{no-visual-clue} problem. For a better illustration, Table \ref{tb_notation} shows some notations used hereinafter.
\begin{table}[h]
	\vspace{-10pt}
	\centering
	\caption{Notation.}
	\setlength{\tabcolsep}{1.8mm}{
		\begin{tabular}{lll}
			\toprule
			Variable  & Type   & Definition                               \\ \midrule
			$H$         & Scalar & Height of image                          \\
			$W$         & Scalar & Width of image                           \\
			$h$         & Scalar & Number of row anchors         \\
			$w$         & Scalar & Number of gridding cells        \\
			$C$         & Scalar & Number of lanes                          \\
			$X$         &Tensor& The global features of image \\
			$f$          &Function& The classifier for selecting lane locations \\
			$P \in R^{C\times h \times (w+1)}$ & Tensor & Group prediction \\ 
			$T \in R^{C\times h \times (w+1)}$ & Tensor & Group target \\ 
			$Prob \in R^{C\times h \times w}$        & Tensor & Probability of each location              \\
			$Loc \in R^{C\times h} $     & Matrix & Locations of lanes                 \\
			\bottomrule
	\end{tabular}}
	\label{tb_notation}
\end{table}


\subsubsection{Definition of our formulation}
\begin{figure*}[t]
	\centering
	\subfigure[Our formulation]{\includegraphics[width=3.0in]{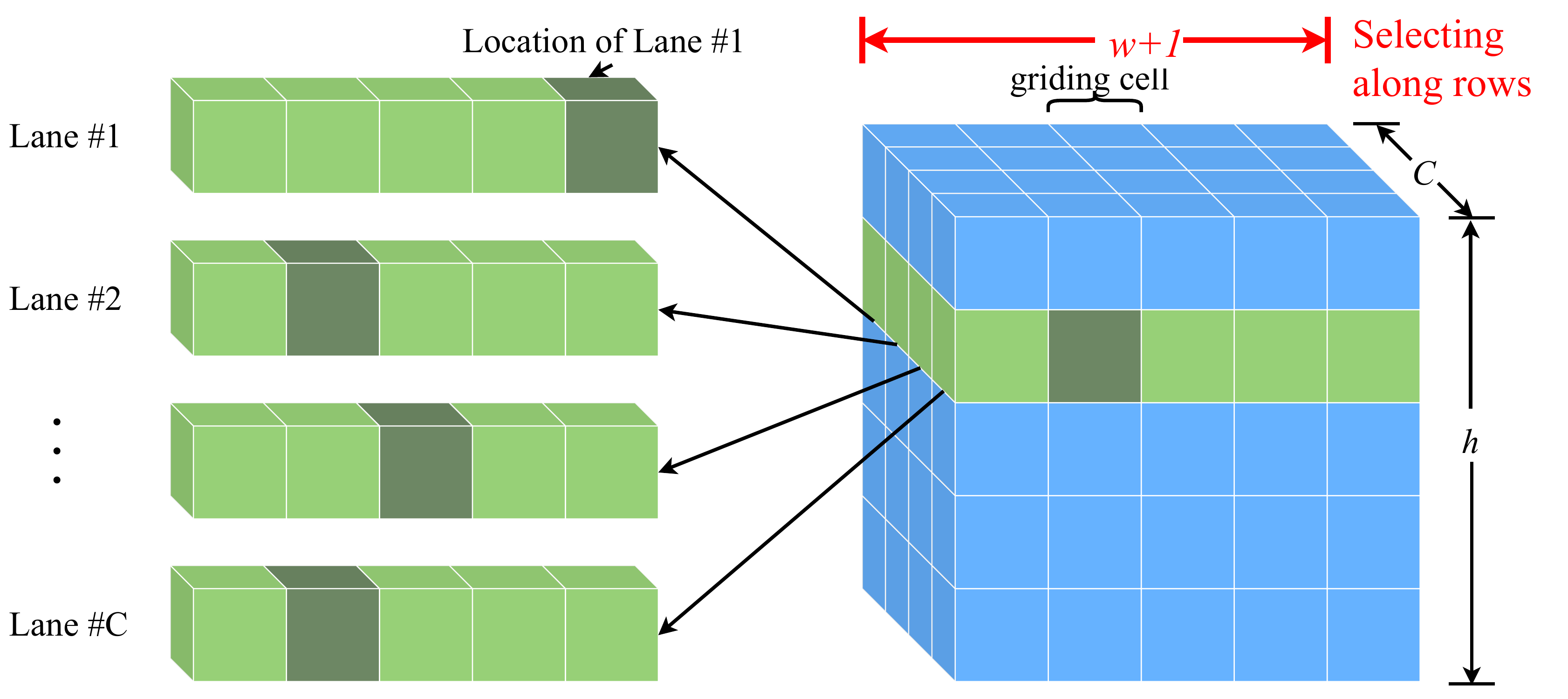}}
	\subfigure[Segmentation]{\includegraphics[width=1.55in]{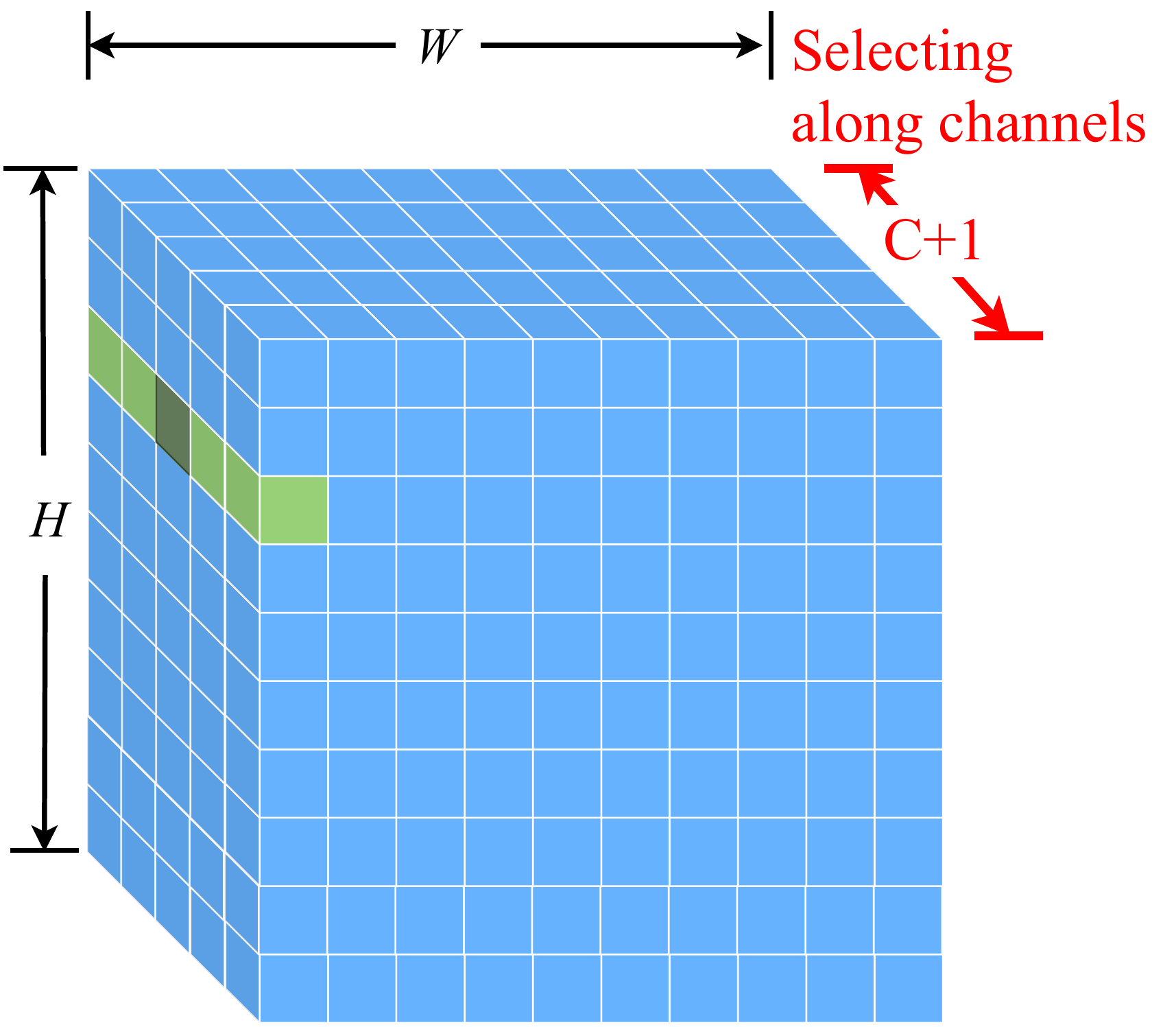}}
	\caption{Illustration of our formulation and conventional segmentation. Our formulation is selecting locations (grids) on rows, while segmentation is classifying every pixel. The dimensions used for classifying are also different, which is marked in red. The proposed formulation significantly reduces the computational cost. Besides, the proposed formulation uses global features as input, which has larger receptive field than segmentation, thus addressing the \textit{no-visual-clue} problem}
	\label{fig_formulation_diff}
	\vspace{-10pt}
\end{figure*}

In order to cope with the problems above, we propose to formulate lane detection to a \textbf{row-based selecting method based on global image features}. In other words, our method is selecting the correct locations of lanes on each predefined row using the global features. In our formulation, lanes are represented as a series of horizontal locations at predefined rows, i.e., row anchors. In order to represent locations, the first step is gridding. On each row anchor, the location is divided into many cells. In this way, the detection of lanes can be described as selecting certain cells over predefined row anchors, as shown in Fig. \ref{fig_formulation_diff}(a). 

Suppose the maximum number of lanes is $C$, the number of row anchors is $h$ and the number of gridding cells is $w$. Suppose $X$ is the global image feature and $f^{ij}$ is the classifier used for selecting the lane location on the $i$-th lane, $j$-th row anchor. Then the prediction of lanes can be written as:
\begin{equation}
	P_{i,j, :} = f^{ij}(X), \ \text{s.t.} \ \  i \in [1,C], j \in [1,h],
	\label{eq_predict}
\end{equation}
in which $P_{i,j,:}$ is the $(w+1)$-dimensional vector represents the probability of selecting $(w+1)$ gridding cells for the $i$-th lane, $j$-th row anchor. Suppose $T_{i,j,:}$ is the one-hot label of correct locations. Then, the optimization of our formulation corresponds to:
\begin{equation}
L_{cls}=\sum_{i=1}^{C} \sum_{j=1}^{h} L_{CE}(P_{i,j, :},T_{i,j, :}),
\end{equation}
in which $L_{CE}$ is the cross entropy loss. 
We use an extra dimension to indicate the absence of lane, so our formulation is composed of $(w+1)$-dimensional instead of $w$-dimensional classifications. 

From Eq. \ref{eq_predict} we can see that our method predicts the probability distribution of all locations on each row anchor based on global features. As a result, the correct location can be selected based on the probability distribution.

\subsubsection{How the formulation achieves fast speed}
The differences between our formulation and segmentation are shown in Fig. \ref{fig_formulation_diff}. It can be seen that our formulation is much simpler than the commonly used segmentation. Suppose the image size is $H \times W$. In general, the number of predefined row anchors and gridding size are far less than the size of an image, that is to say, $h \ll H$ and $w \ll W$. In this way, the original segmentation formulation needs to conduct $H \times W$ classifications that are $(C+1)$-dimensional, while our formulation only needs to solve $C \times h$ classification problems that are $(w+1)$-dimensional. In this way, the scale of computation can be reduced considerably because the computational cost of our formulation is $C\times h\times (w+1)$ while the one for segmentation is $H\times W \times (C+1)$. For example, using the common settings of the CULane dataset \cite{SCNN}, the ideal computational cost of our method is $1.7\times10^4$ calculations and the one for segmentation is $1.15\times 10^6$ calculations. The computational cost is significantly reduced and our formulation could achieve extremely fast speed.

\subsubsection{How the formulation handles the no-visual-clue problem}
In order to handle the \textit{no-visual-clue} problem, utilizing information from other locations is important because \textit{no-visual-clue} means no information at the target location. For example, a lane is occluded by a car, but we could still locate the lane by information from other lanes, road shape, and even car direction. In this way, utilizing information from other locations is key to solve the \textit{no-visual-clue} problem, as shown in Fig. \ref{figure_intro}.

From the perspective of the receptive field, our formulation has a receptive field of the whole image, which is much bigger than segmentation methods. The context information and messages from other locations of the image can be utilized to address the \textit{no-visual-clue} problem. 
From the perspective of learning, prior information like shape and direction of lanes can also be learned using structural loss based on our formulation, as shown in Sec. \ref{sec_loss}. In this way, the \textit{no-visual-clue} problem can be handled in our formulation.
 
Another significant benefit is that this kind of formulation models lane location in a row-based fashion, which gives us the opportunity to establish the relations between different rows explicitly. The original semantic gap, which is caused by low-level pixel-wise modeling and high-level long line structure of lane, can be bridged.

\subsection{Lane structural loss}
\label{sec_loss}

Besides the classification loss, we further propose two loss functions which aim at modeling location relations of lane points. In this way, the learning of structural information can be encouraged.

The first one is derived from the fact that lanes are continuous, that is to say, the lane points in adjacent row anchors should be close to each other. In our formulation, the location of the lane is represented by a classification vector. So the continuous property is realized by constraining the distribution of classification vectors over adjacent row anchors. In this way, the similarity loss function can be:
\begin{equation}
L_{sim}=\sum_{i=1}^{C} \sum_{j=1}^{h-1} \left \| P_{i,j,:} - P_{i,j+1,:} \right \|_1,
\label{eq_sim}
\end{equation}
in which $P_{i,j,:}$ is the prediction on the $j$-th row anchor and $\left \| \cdot \right \|_1$ represents L$_1$ norm.

Another structural loss function focuses on the shape of lanes. Generally speaking, most of the lanes are straight. Even for the curve lane, the majority of it is still straight due to the perspective effect. In this work, we use the second-order difference equation to constrain the shape of the lane, which is zero for the straight case.

To consider the shape, the location of the lane on each row anchor needs to be calculated. The intuitive idea is to obtain locations from the classification prediction by finding the maximum response peak. For any lane index $i$ and row anchor index $j$, the location $Loc_{i,j}$ can be represented as:
\begin{equation}
Loc_{i,j} = \argmax_k \, P_{i,j,k} \,, \  \text{s.t.} \ \   k \in [1,w]  
\label{eq_argmax}
\end{equation}
in which $k$ is an integer representing the location index. It should be noted that we do not count in the background gridding cell and the location index $k$ only ranges from 1 to $w$, instead of $w+1$. 

However, the $argmax$ function is not differentiable and can not be used with further constraints. Besides, in the classification formulation, classes have no apparent order and are hard to set up relations between different row anchors. To solve this problem, we propose to use the expectation of predictions as an approximation of location. We use the softmax function to get the probability of different locations:
\begin{equation}
Prob_{i,j,:} = softmax(P_{i,j,1:w}),
\end{equation}
in which $P_{i,j,1:w}$ is a $w$-dimensional vector and $Prob_{i,j,:}$ represents the probability at each location. For the same reason as Eq. \ref{eq_argmax}, background gridding cell is not included and the calculation only ranges from 1 to $w$. Then, the expectation of locations can be written as: 

\begin{equation}
Loc_{i,j} = \sum_{k=1}^w k \cdot Prob_{i,j,k} 
\label{eq_approx}
\end{equation}
in which $Prob_{i,j,k}$ is the probability of the $i$-th lane, the $j$-th row anchor, and the $k$-th location. The benefits of this localization method are twofold. The first one is that the expectation function is differentiable. The other is that this operation recovers the continuous location with the discrete random variable.

According to Eq. \ref{eq_approx}, the second-order difference constraint can be written as:
\begin{equation}
\begin{split}
L_{shp} = \sum_{i=1}^C \sum_{j=1}^{h-2} & \| (Loc_{i,j} - Loc_{i,j+1}) \\
& - (Loc_{i,j+1} - Loc_{i,j+2}) \|_1,
\end{split}
\end{equation}
in which $Loc_{i,j}$ is the location on the $i$-th lane, the $j$-th row anchor.
The reason why we use the second-order difference instead of the first-order difference is that the first-order difference is not zero in most cases. So the network needs extra parameters to learn the distribution of the first-order difference of lane location. Moreover, the constraint of the second-order difference is relatively weaker than that of the first-order difference, thus resulting in less influence when the lane is not straight. Finally, the overall structural loss can be:
\begin{equation}
L_{str} = L_{sim} + \lambda L_{shp},
\label{eq_str}
\end{equation}
in which $\lambda$ is the loss coefficient.
\subsection{Feature aggregation}
\begin{figure*}[t]
	\centering
	\includegraphics[width=\linewidth]{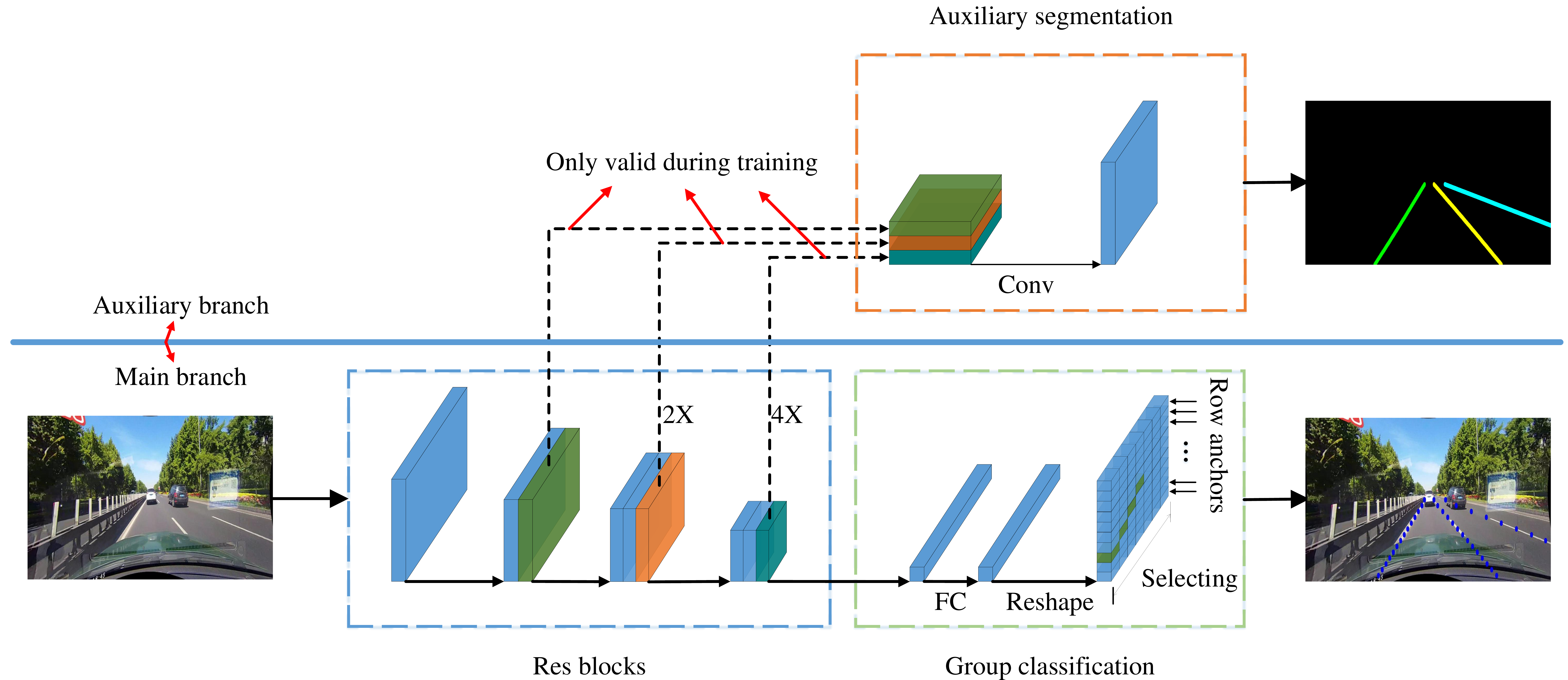}
	\caption{Overall architecture. The auxiliary branch is shown in the upper part, which is only valid when training. The feature extractor is shown in the blue box. The classification-based prediction and auxiliary segmentation task are illustrated in the green and orange boxes, respectively. The group classification is conducted on each row anchor.}
	\label{fig_aggregation}
	\vspace{-10pt}
\end{figure*}

In Sec. \ref{sec_loss}, the loss design mainly focuses on the intra-relations of lanes. In this section, we propose an auxiliary feature aggregation method that performs on the global context and local features. An auxiliary segmentation task utilizing multi-scale features is proposed to model local features. We use cross entropy as our auxiliary segmentation loss. In this way, the overall loss of our method can be written as:
\begin{equation}
L_{total} = L_{cls} + \alpha L_{str} + \beta L_{seg},
\label{eq_total}
\end{equation}
in which $L_{seg}$ is the segmentation loss, $\alpha$ and $\beta$ are loss coefficients.
The overall architecture can be seen in Fig. \ref{fig_aggregation}.

It should be noted that our method only uses the auxiliary segmentation task in the training phase, and it would be removed in the testing phase. In this way, even we added the extra segmentation task, the running speed of our method would not be affected. It is the same as the network without the auxiliary segmentation task.

\section{Experiments}

In this section, we demonstrate the effectiveness of our method with extensive experiments. The following sections mainly focus on three aspects: 1) Experimental settings. 2) Ablation studies of our method. 3) Results on two major lane detection datasets. 

\begin{table*}[]
	\centering
	\caption{Datasets description}
	\setlength{\tabcolsep}{0.6mm}{
		\begin{tabular}{@{}ccccccccc@{}}
			\toprule
			Dataset  & \#Frame & Train  & Validation & Test   & Resolution   & \#Lane & \#Scenarios & environment \\ \midrule
			TuSimple & 6,408   & 3,268  & 358   & 2,782  & 1280$\times$720 & $\leq$5 & 1 & highway     \\
			CULane   & 133,235 & 88,880 & 9,675 & 34,680 & 1640$\times$590 & $\leq$4 & 9 & urban and highway    \\ \bottomrule
	\end{tabular}}
	\label{tab:dataset}
	\vspace{-10pt}
\end{table*}

\subsection{Experimental setting}
\label{sec_setting}

\noindent
\textbf{Datasets.} To evaluate our approach, we conduct experiments on two widely used benchmark datasets: TuSimple Lane detection benchmark \cite{tusimple} and CULane dataset \cite{SCNN}. TuSimple dataset is collected with stable lighting conditions in highways. CULane dataset consists of nine different scenarios, including normal, crowd, curve, dazzle light, night, no line, shadow, and arrow in the urban area. The detailed information about the datasets can be seen in Table \ref{tab:dataset}.

\noindent
\textbf{Evaluation metrics.} The official evaluation metrics of the two datasets are different. For TuSimple dataset, the main evaluation metric is accuracy. The accuracy is calculated by: 
\begin{equation}
accuracy = \dfrac{\sum_{clip}C_{clip}}{\sum_{clip}S_{clip}} ,
\end{equation}
in which $C_{clip}$ is the number of lane points predicted correctly and $S_{clip}$ is the total number of ground truth in each clip.
As for the evaluation metric of CULane, each lane is treated as a 30-pixel-width line. Then the intersection-over-union (IoU) is computed between ground truth and predictions. Predictions with IoUs larger than 0.5 are considered as true positives. F1-measure is taken as the evaluation metric and formulated as follows:
\begin{equation}
F1-measure = \frac{2 \times Precision \times Recall}{Precision + Recall},
\end{equation}
where $Precision = \frac{TP}{TP + FP}$, $Recall = \frac{TP}{TP + FN}$ , $TP$ is the true positive, $FP$ is the false positive, and $FN$ is the false negative.

\noindent
\textbf{Implementation details.} For both datasets, we use the row anchors that are defined by the dataset. Specifically, the row anchors of Tusimple dataset, in which the image height is 720, range from 160 to 710 with a step of 10. The counterpart of CULane dataset ranges from 260 to 530, with the same step as Tusimple. The image height of CULane dataset is 540. The number of gridding cells is set to 100 on the Tusimple dataset and 150 on the CULane dataset. The corresponding ablation study on the Tusimple dataset can be seen in Sec. \ref{ab_griding}. 

In the optimizing process, images are resized to 288$\times$800 following \cite{SCNN}. We use Adam \cite{kingma2014adam} to train our model with cosine decay learning rate strategy \cite{loshchilov2016sgdr} initialized with 4e-4. Loss coefficients $\lambda$, $\alpha$ and $\beta$ in Eq. \ref{eq_str} and \ref{eq_total} are all set to 1. The batch size is set to 32, and the total number of training epochs is set 100 for TuSimple dataset and 50 for CULane dataset. The reason why we choose such a large number of epochs is that our structure-preserving data augmentation requires a long time of learning. The details of our data augmentation method are discussed in what follows. All models are trained and tested with pytorch \cite{paszke2017automatic} and nvidia GTX 1080Ti GPU. 



\begin{figure}
	\centering
	\subfigure[Original anaotation]{\includegraphics[width=2in]{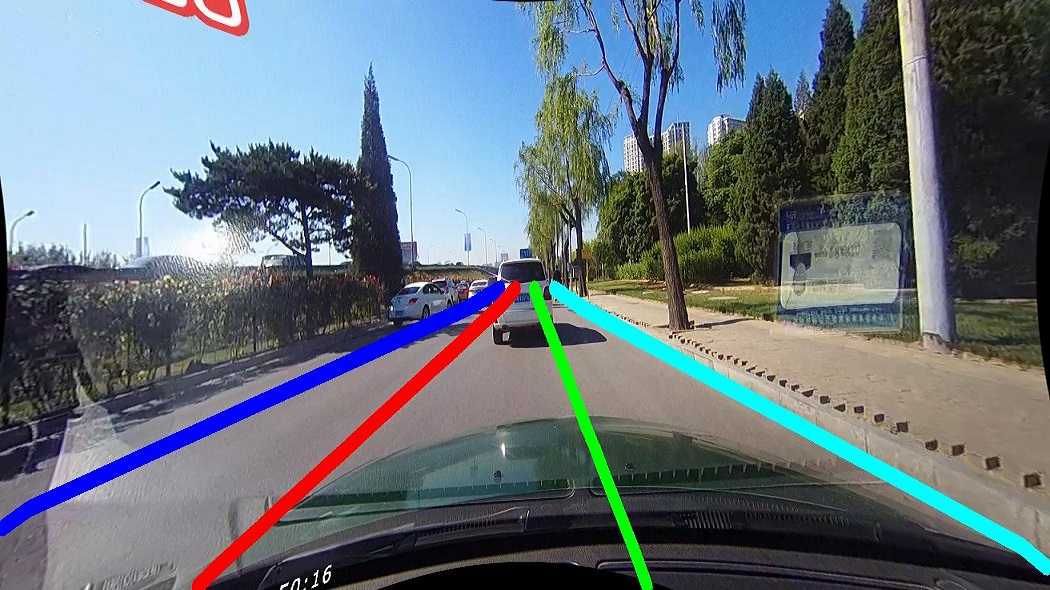}}
	\subfigure[Augmentated result]{\includegraphics[width=2in]{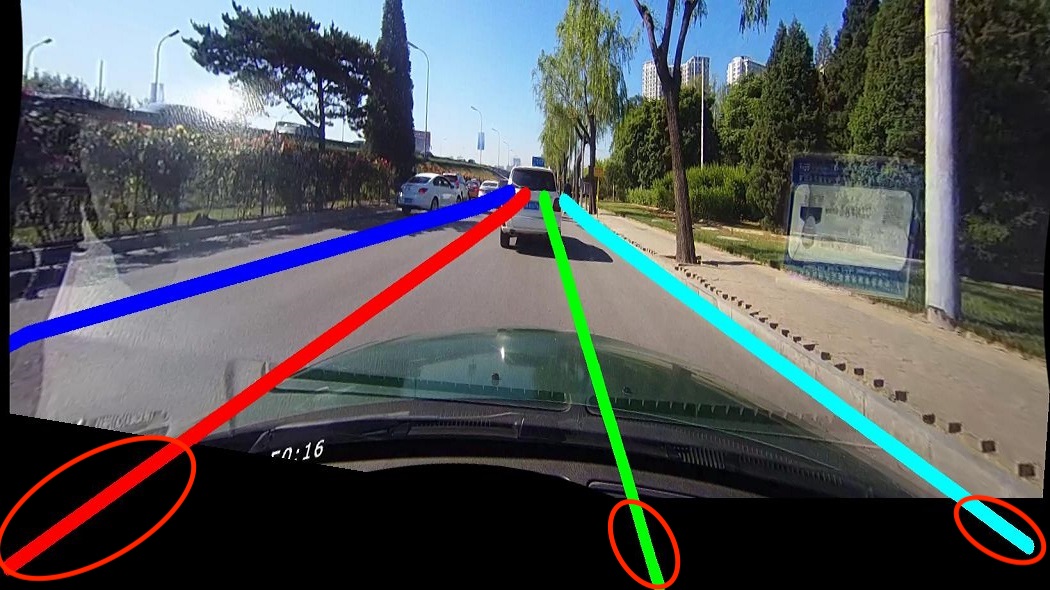}}
	\caption{Demonstration of augmentation. The lane on the right image is extended to maintain the lane structure, which is marked with red ellipse.}
	\label{fig_aug}
	\vspace{-10pt}
\end{figure}

\noindent
\textbf{Data augmentation.}
Due to the inherent structure of lanes, a classification-based network could easily over-fit the training set and show poor performance on the validation set. To prevent this phenomenon and gain generalization ability, an augmentation method composed of rotation, vertical and horizontal shift is utilized. Besides, in order to preserve the lane structure, the lane is extended till the boundary of the image. The results of augmentation can be seen in Fig. \ref{fig_aug}.

\subsection{Ablation study}
In this section, we verify our method with several ablation studies. The experiments are all conducted with the same settings as Sec. \ref{sec_setting}.

\begin{figure}[h]
	\centering
	\includegraphics[width=0.45\columnwidth]{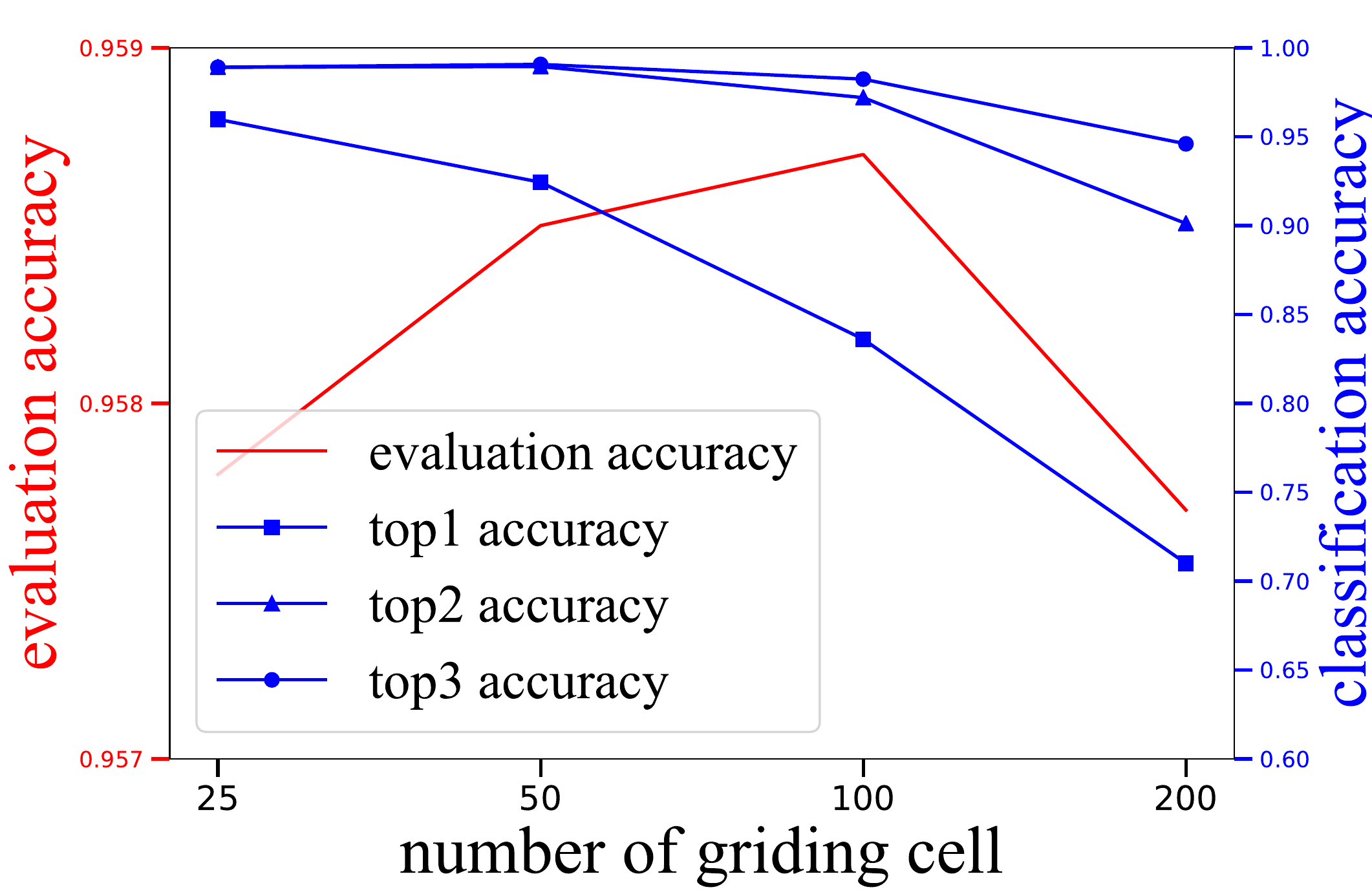}
	\caption{Performance under different numbers of gridding cells on the Tusimple Dataset. Evaluation accuracy means the evaluation metric proposed in the Tusimple benchmark, while classification accuracy is the standard accuracy. Top1, top2 and top3 accuracy are the metrics when the distance of prediction and ground truth is less than 1, 2 and 3, respectively. In this figure, top1 accuracy is equivalent to standard classification accuracy.}
	\label{fig_griding}
	\vspace{-10pt}
\end{figure}

\noindent
\textbf{Effects of number of gridding cells.}
\label{ab_griding}
As described in Sec. \ref{sec_formulation}, we use gridding and selecting to establish the relations between structural information in lanes and classification-based formulation. In this way, we further try our method with different numbers of gridding cells to demonstrate the effects on our method. We divide the image using 25, 50, 100 and 200 cells in columns. The results can be seen in Figure. \ref{fig_griding}. 

With the increase of the number of gridding cells, we can see that both top1, top2 and top3 classification accuracy drops gradually. It is because more gridding cells require finer-grained and harder classification. However, the evaluation accuracy is not strictly monotonic. Although a smaller number of gridding cells means higher classification accuracy, the localization error would be larger, since the gridding cell is too large to represent precise location. In this work, we choose 100 as our number of gridding cells on the Tusimple Dataset.

\noindent
\textbf{Effectiveness of localization methods.}
Since our method formulates the lane detection as a group classification problem, one natural question is what are the differences between classification and regression. In order to test in a regression manner, we replaced the group classification head in Fig. \ref{fig_aggregation} with a similar regression head. We use four experimental settings, which are respectively REG, REG Norm, CLS and CLS Exp. CLS means the classification-based method, while REG means the regression-based method. The difference between CLS and CLS Exp is that their localization methods are different, which are respectively Eq. \ref{eq_argmax} and Eq. \ref{eq_approx}. The REG Norm setting is the variant of REG, which normalizes the learning target.

\begin{table}[h]
	\vspace{-10pt}
	\centering
	\caption{Comparison between classification and regression on the Tusimple dataset. REG and REG Norm are regression-based methods, while the ground truth scale of REG Norm is normalized. CLS means standard classification with the localization method in Eq. \ref{eq_argmax} and CLS Exp means the one with Eq. \ref{eq_approx}. }
	\setlength{\tabcolsep}{3.0mm}{
	\begin{tabular}{ccccc}
	\toprule
	Type     & REG & REG Norm & CLS & CLS Exp \\ \midrule
	Accuracy      & 71.59    &67.24 &95.77 & 95.87         \\ \bottomrule
	
\end{tabular}}
	\label{tb_cls_reg}
	\vspace{-10pt}
\end{table}

The results are shown in Table \ref{tb_cls_reg}. We can see that classification with the expectation could gain better performance than the standard method. This result also proves the analysis in Eq. \ref{eq_approx} that the expectation based localization is more precise than $argmax$ operation. Meanwhile, classification-based methods could consistently outperform the regression-based methods.

\noindent
\textbf{Effectiveness of the proposed modules.} In order to verify the effectiveness of the proposed modules, we conduct both qualitative and quantitative experiments. 

First, we show the quantitative results of our modules. As shown in Table \ref{tab:Module ablation}, the experiments are carried out with the same training settings and different module combinations.

\begin{table}[h]
	\centering
	\caption{Experiments of the proposed modules on Tusimple benchmark with Resnet-34 backbone. Baseline stands for conventional segmentation formulation.}
	\label{tab:Module ablation}
	\setlength{\tabcolsep}{1.4mm}{
		\begin{tabular}{ccccc}
			\toprule
			Baseline & New formulation& Structural loss & Feature aggregation & Accuracy \\ \midrule
			\checkmark & & & & 92.84 \\
			 & \checkmark &  & & 95.64(+2.80)  \\
			 & \checkmark& \checkmark &  & 95.96(+3.12) \\
			 & \checkmark &  & \checkmark & 95.98(+3.14) \\
			 & \checkmark & \checkmark & \checkmark & 96.06(+3.22) \\ \bottomrule
	\end{tabular}}
\end{table}

From Table \ref{tab:Module ablation}, we can see that the new formulation gains significant performance improvement compared with segmentation formulation. Besides, both lane structural loss and feature aggregation could enhance the performance, which proves the effectiveness of the proposed modules. 

\begin{figure}[h]
	
	\centering
	\subfigure[W/O similarity loss]{\includegraphics[width=2.2in]{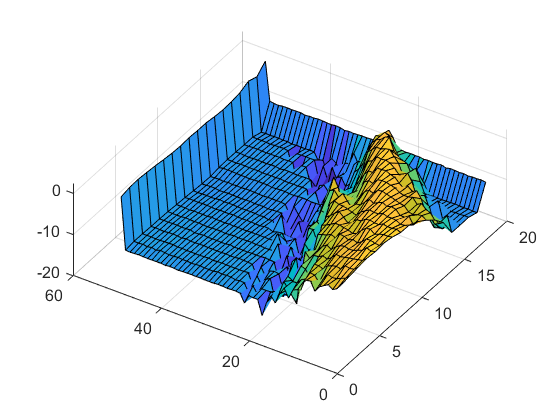}}
	\subfigure[W/ similarity loss]{\includegraphics[width=2.2in]{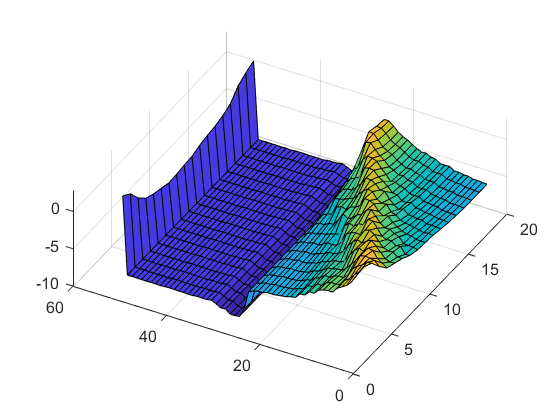}}
	\caption{Qualitative comparison of similarity loss. The predicted distributions of group classification of the same lane are shown. Fig. (a) shows the visualization of distribution without similarity loss, while Fig. (b) shows the counterpart with similarity loss.}
	\label{fig_col_test}
	\vspace{-10pt}
\end{figure}

Second, we illustrate the effectiveness of lane similarity loss in Eq. \ref{eq_sim}. The results are shown in Fig. \ref{fig_col_test}. We can see that similarity loss makes the classification prediction smoother and thus gains better performance.

\subsection{Results}
In this section, we show the results on two lane detection datasets, which are the Tusimple lane detection benchmark and the CULane dataset. In these experiments, Resnet-18 and Resnet-34 \cite{resnet} are used as our backbone models.

For the Tusimple lane detection benchmark, seven methods are used for comparison, including Res18-Seg \cite{chen2017deeplab}, Res34-Seg \cite{chen2017deeplab}, LaneNet \cite{End-to-End}, EL-GAN \cite{ghafoorian2018gan}, SCNN \cite{SCNN} and SAD \cite{SAD}. Both Tusimple evaluation accuracy and runtime are compared in this experiment. The runtime of our method is recorded with the average time for 100 runs. The results are shown in Table \ref{tab:Tusimple compare}.

From Table \ref{tab:Tusimple compare}, we can see that our method achieves comparable performance with state-of-the-art methods while our method could run extremely fast. The biggest runtime gap between our method and SCNN is that our method could infer 41.7 times faster. Even compared with the second-fastest network SAD, our method is still more than 2 times faster. 

\begin{table}[h]
	\vspace{-10pt}
	\centering
	\caption{Comparison with other methods on TuSimple test set. The calculation of runtime multiple is based on the slowest method SCNN.}
	\label{tab:Tusimple compare}
	\begin{tabular}{lccc}
		\toprule
		Method  & Accuracy & Runtime(ms) & Multiple \\ \midrule
		Res18-Seg \cite{chen2017deeplab}& 92.69    & 25.3        &   5.3x       \\
		Res34-Seg \cite{chen2017deeplab} & 92.84    & 50.5        &   2.6x      \\
		LaneNet \cite{End-to-End}  & 96.38    & 19.0        &   7.0x      \\
		EL-GAN \cite{ghafoorian2018gan}    & 96.39    & $>$100        &   $<$1.3x   \\
		SCNN \cite{SCNN}    & 96.53    & 133.5       &   1.0x       \\
		SAD \cite{SAD}      & \textbf{96.64}    & 13.4        &   10.0x     \\
		\midrule
		Res34-Ours      & 96.06    & 5.9        &   22.6x     \\ 
		Res18-Ours      & 95.87    & \textbf{3.2}         &   \textbf{41.7x}     \\ 
		
		\bottomrule
	\end{tabular}
	\vspace{-10pt}
\end{table}

Another interesting phenomenon we should notice is that our method gains both better performance and faster speed when the backbone network is the same as plain segmentation. This phenomenon shows that our method is better than the plain segmentation and verifies the effectiveness of our formulation.

For the CULane dataset, four methods, including Seg\cite{chen2017deeplab}, SCNN \cite{SCNN}, FastDraw \cite{FastDraw} and SAD \cite{SAD}, are used for comparison. F1-measure and runtime are compared. The runtime of our method is also recorded with the average time for 100 runs. The results can be seen in Table \ref{tab:CULane compare}. 
\begin{table*}[]
	\vspace{-15pt}
	\centering
	\caption{Comparison of F1-measure and runtime on CULane testing set with IoU threshold=0.5. For crossroad, only false positives are shown. The less, the better. `-' means the result is not available.}
	\label{tab:CULane compare}
	\setlength{\tabcolsep}{0.4mm}{
		\begin{tabular}{lcccccccc}
			\toprule
			Category     & Res50-Seg\cite{chen2017deeplab} & SCNN\cite{SCNN}  & FD-50\cite{FastDraw} & Res34-SAD & SAD\cite{SAD}  & Res18-Ours & Res34-Ours \\ \midrule
			Normal       & 87.4      & 90.6  & 85.9     & 89.9    & 90.1   & 87.7    & \textbf{90.7}  \\
			Crowded      & 64.1      & 69.7  & 63.6     & 68.5    & 68.8   & 66.0& \textbf{70.2}  \\
			Night        & 60.6      & 66.1  & 57.8     & 64.6    & 66.0   & 62.1& \textbf{66.7}  \\
			No-line      & 38.1      & 43.4  & 40.6     & 42.2    & 41.6   & 40.2& \textbf{44.4}  \\
			Shadow       & 60.7      & 66.9  & 59.9     & 67.7    & 65.9   & 62.8& \textbf{69.3}  \\
			Arrow        & 79.0      & 84.1  & 79.4     & 83.8    & 84.0   & 81.0& \textbf{85.7}  \\
			Dazzlelight & 54.1      & 58.5  & 57.0     & 59.9    & \textbf{60.2}   & 58.4& 59.5  \\
			Curve        & 59.8      & 64.4  & 65.2     & 66.0    & 65.7   & 57.9& \textbf{69.5}  \\
			Crossroad    & 2505      & 1990  & 7013     & 1960    & 1998   & \textbf{1743}& 2037  \\ \midrule
			Total        & 66.7      & 71.6  &  -       & 70.7    & 70.8   & 68.4& \textbf{72.3}  \\ \midrule
			Runtime(ms)  &  -        & 133.5 &  -       & 50.5    & 13.4   & 3.1& 5.7   \\ 
			Multiple     &  -        & 1.0x  &  -       &  2.6x   &  10.0x & {43.0x}  & 23.4x \\   
			FPS          &  -        &  7.5  &  -       & 19.8    &  74.6  & {322.5} & 175.4 \\
			\bottomrule
	\end{tabular}}
	\vspace{-10pt}
\end{table*}

It is observed in Table \ref{tab:CULane compare} that our method achieves the best performance  in terms of both accuracy and speed. It proves the effectiveness of the proposed formulation and structural loss on these challenging scenarios because our method could utilize global and structural information to address the \textit{no-visual-clue} and speed problem. The fastest model of our formulation achieves 322.5 FPS with a resolution of 288$\times$800, which is the same as other compared methods.

\begin{figure*}[t]

	\centering
	\includegraphics[width=0.9\linewidth,height=3.2in]{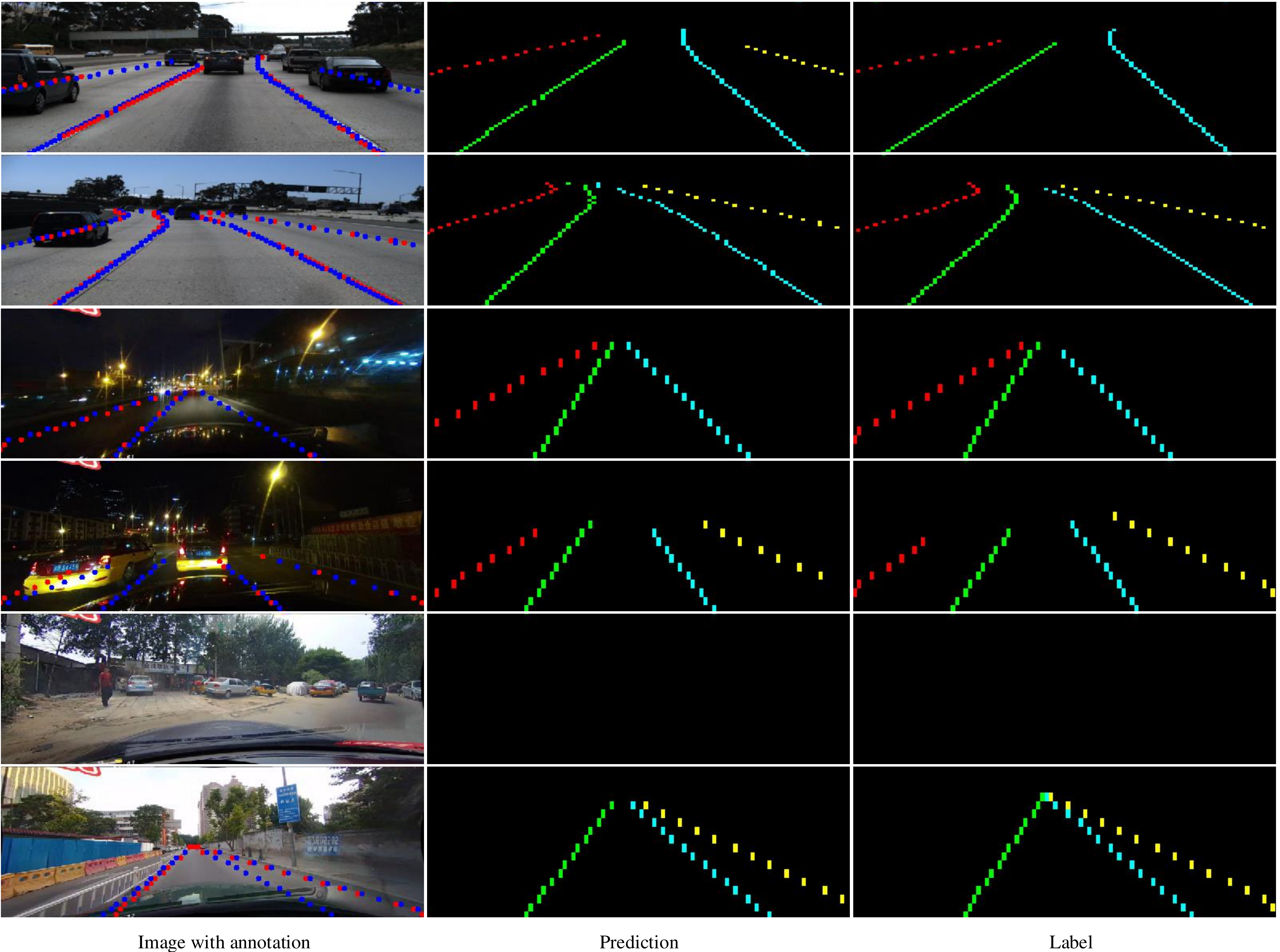}
	\caption{Visualization on the Tusimple and the CULane dataset. The first two rows are results on the Tusimple dataset and the rest rows are results on the CULane dataset. From left to right, the results are image, prediction and label. In the image, predictions are marked in blue and ground truth are marked in red. Because our method only predicts on the predefined row anchors, the scales of images and labels in the vertical direction are not identical.}
	\label{fig_vis}
	\vspace{-15pt}
\end{figure*}

The visualizations of our method on the Tusimple and CULane datasets are shown in Fig. \ref{fig_vis}. We can see our method performs well under various conditions. 
\vspace{-5pt}

\section{Conclusion}


In this paper, we have proposed a novel formulation with structural loss and achieves remarkable speed and accuracy. The proposed formulation regards lane detection as a problem of row-based selecting using global features. In this way, the problem of speed and \textit{no-visual-clue} can be addressed. Besides, structural loss used for explicitly modeling of lane prior information is also proposed. The effectiveness of our formulation and structural loss are well justified with both qualitative and quantitative experiments. Especially, our model using Resnet-34 backbone could achieve state-of-the-art accuracy and speed. A light weight Resnet-18 version of our method could even achieve 322.5 FPS with a comparable performance at the same resolution.

\clearpage

%
%
\bibliographystyle{splncs04}
\bibliography{egbib}
\end{document}